\documentclass[conference]{IEEEtran}
\IEEEoverridecommandlockouts

\usepackage{cite}
\usepackage{amsmath,amssymb,amsfonts}
\usepackage{graphicx}
\usepackage{textcomp}
\usepackage{newtxtext} 
\usepackage[table]{xcolor}  
\usepackage{booktabs}
\usepackage{multirow}
\usepackage{adjustbox}
\usepackage{url}
\usepackage[most]{tcolorbox}
\usepackage{stfloats}   
\usepackage{placeins}   
\usepackage{float}      

\definecolor{rowTrain}{RGB}{218,232,247}  
\definecolor{rowEval}{RGB}{223,240,216}   
\definecolor{rowEmo}{RGB}{252,232,210}    
\definecolor{colFT}{RGB}{223,235,247}     

\begin{document}

\title{OmniTag-LLM: Multilingual Multitask Speech Tagging}

\author{%
\IEEEauthorblockN{%
Aref Farhadipour\textsuperscript{1,2},
Francesco Verdini\textsuperscript{1,3},
Antonis Asonitis\textsuperscript{1,4},\\
Pierre-Edouard Honnet\textsuperscript{1},
Marzieh Razavi\textsuperscript{1},
Vijeta Avijeet\textsuperscript{1},
Juan Pablo Zuluaga Gomez\textsuperscript{1}%
}
\IEEEauthorblockA{%
\textsuperscript{1}AGIGO,
\textsuperscript{2}University of Zurich,
\textsuperscript{3}University of Rome,
\textsuperscript{4}ETH Zurich%
}
}

\maketitle

\begin{abstract}

Multimodal LLMs promise to replace the traditional cascade of specialist speech models with a single system that, from one audio clip, jointly transcribes and labels emotion, gender, language, and named entities. Yet most benchmarks still score these tasks in isolation, and largely in English. We introduce OmniTag-Bench, a five-language benchmark (en/de/es/fr/it) and training recipe for joint understanding: one audio clip in, one JSON object out (transcript, emotion, gender, LID, token-level NER). We evaluate ten open multimodal LLMs and a LoRA fine-tuned Qwen3-Omni on a unified prompt and scorer. Beyond aggregate scores, OmniTag-Bench reveals a systematic failure mode: several strong models often transcribe or translate non-English speech into English, inflating multilingual WER. Fine-tuning improves the base model on NER, emotion, and WER while nearly doubling accuracy on independent human-labeled emotion corpora, suggesting transferable gains beyond pseudo-label imitation. The fine-tuned model also generalizes to unseen but related languages, transferring to held-out Dutch. We release prompts, the scorer, and the labeling pipeline.
\end{abstract}


\section{Introduction}

\begin{figure}[ht]
\centering
\includegraphics[
    trim=1cm 18cm 1.2cm 0.5cm, 
    clip,
    width=0.5\textwidth,
    keepaspectratio
]{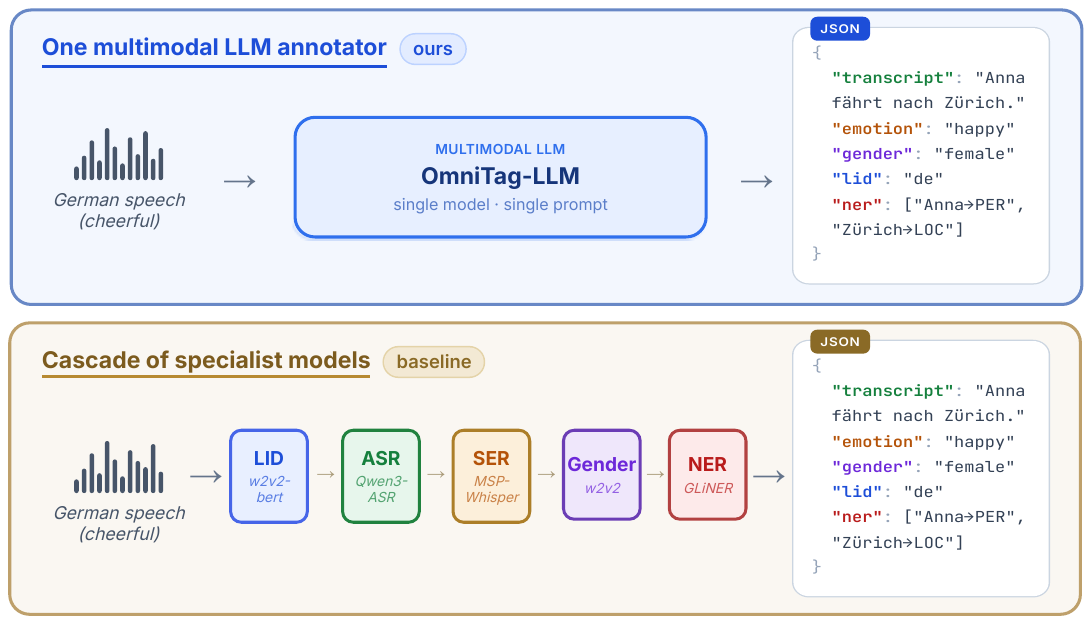}

\caption{A single multimodal LLM (a) jointly generates transcript, emotion,
gender, language identification (LID), and named-entity recognition (NER) in one
pass, versus a conventional pipeline (b) of five specialized models. The German
example illustrates language-agnostic annotation.}

\label{fig:multimodal}
\end{figure}

Multimodal large language models (MLLMs) are emerging as general-purpose
interfaces for audio understanding, where a single model, given one audio clip
and one prompt, can emit labels for several complementary tasks that a
traditional pipeline distributes across dedicated expert
systems~\cite{chu2024qwen2audio,chu2023qwenaudio,deshmukh2023pengi,gong2024ltu,xu2025qwen25omni, farhadipour2025multimodal, farhadipour2026towards, tang2024salmonn,farhadipour2024comparative}
(Fig.~\ref{fig:multimodal}). This consolidation is appealing, yet these
generalists still trail purpose-built specialists on each individual task. A
rapidly expanding ecosystem of open-weight omni and speech LLMs, spanning four to
thirty billion
parameters~\cite{xu2025qwen25omni,qwen3omni,microsoft2025phi4,ding2025kimi,minicpmo2024,mossaudio,stepaudio2,gemma,nemotronomni},
is narrowing this gap, making it timely to measure how closely one model can
approach specialist-level performance across the full annotation stack.

Yet evaluation has not kept pace. Instruction-following audio suites such as
AIR-Bench~\cite{yang2024airbench}, Dynamic-SUPERB~\cite{huang2024dynamicsuperb},
and AudioBench~\cite{wang2024audiobench} broadened testing beyond single tasks;
ISA-Bench~\cite{li2025isabench} probed instruction sensitivity, and
MMAU~\cite{mmau} extended coverage to speech, sound, and music reasoning. These
suites nonetheless largely (i)~grade one question per query rather than all tasks
jointly; (ii)~target English, confining multilingual evaluation mainly to ASR and
language identification (ML-SUPERB~\cite{mlsuperb} scales to $100+$ languages but
scores encoder representations, not MLLM generation); and (iii)~accept free-form
answers that are hard to grade reliably at scale. One competency thus remains
unmeasured, namely whether a single model can deliver \emph{joint, structured,
multilingual} output from one prompt for multilingual input. OmniTag-Bench fixes the instruction but
demands \emph{all five} tasks (ASR, emotion, gender, language identification
(LID), and named-entity recognition (NER)) at once across \emph{five} languages
(English, German, Spanish, French, Italian) as a single JSON object, stressing
format compliance and cross-lingual consistency rather than instruction variety.

We close this gap with a benchmark, a model, and a broad empirical study.
High-quality manual labels for emotion, gender, and NER are costly to obtain at
scale, so, following common practice, we distill them offline from specialist
teachers, a Whisper-based classifier~\cite{radford2023whisper} for emotion, a
lightweight audio classifier~\cite{baevski2020wav2vec2} for gender, and
GLiNER2~\cite{gliner2} for NER. These distilled labels both supervise the joint
model and backfill those missing from individual corpora, whereas transcripts and
language remain native throughout. The pipeline itself is dataset-agnostic, so given any audio
collection it can produce training-ready labels for all five tasks, falling back
to a dedicated specialist wherever native annotations are unavailable. For
transcription it uses Qwen3-ASR-1.7B~\cite{shi2026qwen3} and for LID a w2v-BERT
language-identification model~\cite{farhadipour2026subtract}, completing the
five-specialist set of Fig.~\ref{fig:multimodal}. We then
evaluate the resulting model against ten open
baselines under one joint, multilingual protocol. Specifically, we contribute:
\begin{enumerate}
\item \textbf{OmniTag-LLM.} A multilingual multitask speech-labeling model built
on Qwen3-Omni. Fine-tuning improves the base model on every task
(Table~\ref{tab:fullresults}), with the most diagnostic gains on
\emph{held-out, human-labeled} emotion corpora independent of the distillation
teachers, evidence of transferable rather than merely imitative competence. The gains also carry over to Dutch, a language outside the five training languages (Section~\ref{sec:generalization}).
\item \textbf{OmniTag-Bench and curated label sets.} Five language splits with
closed-vocabulary JSON targets for ASR, emotion, gender, LID, and NER (NER from
MSNER~\cite{meeus2024msner}; all labels native except teacher-pseudo-labeled
emotion), evaluated under an identical prompt and scorer for all systems.
\item \textbf{A broad model survey.} Ten open-weight MLLMs spanning
$4$B--$30$B parameters and four serving back-ends, under one joint decoding
protocol.
\item \textbf{A cross-lingual failure analysis.} Identification, quantification,
and partial mitigation of a pervasive ``transcribe-into-English'' failure mode,
with practical implications for multilingual MLLM reporting.
\end{enumerate}

\section{The OmniTag-Bench Evaluation Suite}
\subsection{Task formulation}
Each example is one audio clip $x$. A fixed instruction asks the model to return
one JSON object $\hat{y}$ with five keys: verbatim \texttt{transcript},
\texttt{emotion}, \texttt{gender}, \texttt{lid}, and \texttt{ner}. The four
non-ASR tasks use closed label vocabularies (Table~\ref{tab:labels}) that appear
identically in the gold targets and the prompt, so a model is always shown the
legal values. The verbatim prompt is given in Fig.~\ref{fig:prompt}; all systems
are graded against the same gold JSON.

\subsection{Evaluation data}\label{sec:evaldata}
OmniTag-Bench is built on VoxPopuli~\cite{voxpopuli}, a large multilingual corpus
of European Parliament recordings, using one split per language (English, German,
Spanish, French, Italian) totalling $8{,}000$ utterances
(Table~\ref{tab:datastats}). Each clip is scored for all five tasks: transcript,
gender, and language labels are native to VoxPopuli; NER comes from
MSNER~\cite{meeus2024msner} (a multilingual speech-NER set built on the same
recordings) for every language except Italian, which MSNER does not cover and
whose NER we therefore pseudo-label with the GLiNER2 teacher; and emotion is
pseudo-labeled by the expert teacher used for training (Section~\ref{sec:experts}). To avoid evaluating emotion only against
pseudo-labels, we additionally report native speech-emotion recognition (SER)~\cite{pepino2021emotion,wagner2023dawn,akcay2020speech} on
one acted set per language, IEMOCAP~\cite{iemocap}, EmoDB~\cite{emodb},
MESD~\cite{mesd}, CaFE~\cite{cafe}, and EMOVO~\cite{emovo}
(Table~\ref{tab:datastats}). On these sets the model receives the same joint
five-task prompt, but we report emotion only, as no ground truth is available for
the other four tasks.

\begin{table}[t]
\caption{The joint output schema at a glance: the five tasks emitted in one JSON
object, their output granularity, and label spaces. ASR is free text; the other
four use closed vocabularies that appear identically in the gold targets and the
prompt.}
\label{tab:labels}
\centering
\footnotesize
\setlength{\tabcolsep}{4pt}
\renewcommand{\arraystretch}{1.15}
\begin{tabular}{@{}llcp{0.49\columnwidth}@{}}
\toprule
\textbf{Task} & \textbf{Output} & \textbf{\#} & \textbf{Label space} \\
\midrule
ASR     & free text   & --- & verbatim transcript, spoken language \\
Emotion & utterance   & 9 & anger, contempt, disgust, fear, happiness, neutral, sadness, surprise, other \\
Gender  & utterance   & 2 & male, female \\
LID     & utterance   & 6 & english, french, spanish, german, italian, other \\
NER     & token (BIO) & 9$\rightarrow$19 & types: DATE, EVENT, GPE, MONEY, NORP, ORG, PERSON, PRODUCT, WORK\_OF\_ART; tags: \texttt{O}, \texttt{B-/I-} per type \\
\bottomrule
\end{tabular}
\end{table}

\subsection{Unified prompt and output schema}
A single preamble instructs the model to ``respond with one JSON object only,''
lists the five tasks as bullet points with their allowed value sets.
The \emph{identical} prompt builds the training targets and queries every evaluated
model, removing train/test prompt mismatch.
The ASR bullet requires transcription \emph{in the spoken language} and forbids translation, a constraint
motivated and analyzed in Section~\ref{sec:crosslingual}.

\begin{figure*}[t]
\centering
\begin{tcolorbox}[enhanced, breakable=false, width=\textwidth,
  colback=blue!2!white, colframe=blue!55!black, boxrule=0.8pt, arc=4pt,
  left=8pt, right=8pt, top=5pt, bottom=5pt,
  fonttitle=\bfseries\small, coltitle=white,
  title={Shared Joint Speech-Understanding Prompt %
  \hfill\normalfont\itshape\footnotesize  (identical for training targets \& evaluation)},
  attach boxed title to top left={xshift=8pt, yshift=-2pt},
  boxed title style={colback=blue!55!black, arc=2pt}]
\small
\textbf{\textsf{System.}} You are a speech understanding model. Listen to the
audio (\texttt{<audio>}) and reply with \textbf{one JSON object only}, with no
markdown fences and no commentary. Perform all five tasks \textbf{jointly}, one key
per task:
\smallskip

\renewcommand{\arraystretch}{1.25}
\begin{tabular}{@{}r@{\ \ }p{0.86\textwidth}@{}}
\textcolor{blue!55!black}{\textbf{ASR}} &
transcribe the audio \emph{verbatim} in the spoken language; \textbf{do not translate} into any other language. Preserve punctuation, filler words, and casing as spoken.\\
\textcolor{blue!55!black}{\textbf{Emotion}} &
exactly one of: \texttt{anger, contempt, disgust, fear, happiness, neutral, sadness, surprise, other}.\\
\textcolor{blue!55!black}{\textbf{NER}} &
token-level BIO tags (\texttt{O}, \texttt{B-}/\texttt{I-}\textsc{type}) over 9 types:
\texttt{DATE, EVENT, GPE, MONEY, NORP, ORG, PERSON, PRODUCT, WORK\_OF\_ART}.\\
\textcolor{blue!55!black}{\textbf{Gender}} &
exactly one of: \texttt{male, female}.\\
\textcolor{blue!55!black}{\textbf{LID}} &
exactly one of: \texttt{english, french, german, italian, spanish, other} (use \texttt{other} if the spoken language is not listed).\\
\end{tabular}

\end{tcolorbox}
\caption{The shared joint-task prompt, shown verbatim. The \emph{same} prompt
builds the supervised training targets and queries every evaluated model, so all
systems see byte-identical instructions.}
\label{fig:prompt}
\end{figure*}

\subsection{Metrics}\label{sec:metrics}
A single scorer is applied uniformly to all models. ASR uses corpus-level word
and character error rate (WER/CER) after Unicode NFKC
normalization, lowercasing, and punctuation removal. Gender and LID are reported
by accuracy and emotion uses accuracy and macro-F1  (Table~\ref{tab:fullresults}). We report both for emotion deliberately. VoxPopuli is heavily biased
toward \emph{neutral}, so accuracy rewards a majority-class predictor and can look
deceptively high, whereas macro-F1 reveals
whether minority affect classes are recovered at all. The balanced
native-emotion corpora (Section~\ref{sec:evaldata}) are our primary test of
emotion competence.

Following standard practice, we score entities rather than tokens. From each
predicted and gold BIO sequence we extract spans as \texttt{(phrase, type)} pairs
and compute micro-averaged precision, recall, and F1 under two matching criteria.
Entity-F1 is the strict CoNLL-2003 standard~\cite{conll2003} and counts a
span as correct only when both its type and its normalized surface form match.
Label-F1 follows the SLUE convention~\cite{slue} and requires only the
type to match, which makes it robust to the spelling and segmentation errors that
ASR introduces. We deliberately avoid the common seqeval ``chunk-F1,'' which
requires the predicted and gold token sequences to have equal length. MLLMs
frequently truncate their BIO lists, so for some models the length-matched subset
falls below $1\%$ of clips and the score becomes unreliable and heavily deflated.
Both of our criteria are instead computed over $100\%$ of clips. We also report
precision and recall alongside F1 because models fail asymmetrically. A model that
under-generates entities scores high precision but low recall, which differs
qualitatively from one that over-predicts them.


\section{Expert-Distilled Training Data}
\label{sec:training}
To fine-tune OmniTag-LLM we merge complementary corpora into one stream under a
\emph{task-aware} schema. A per-example flag records which labels each clip
carries, so corpora with disjoint coverage combine cleanly~\cite{farhadipour2025adaptive} and a single clip can
supervise several tasks. Because no public corpus provides all five labels
jointly, we fill each corpus's missing labels with task-specific \emph{expert
teachers} (Section~\ref{sec:experts}). Transcripts and language identity are
never machine-generated, taken from native references after markup cleanup and
from native metadata.

\subsection{Source corpora}
The training mixture has two parts (Table~\ref{tab:datastats}). Four English
spoken-NER corpora~\cite{ghannay2018end,yadav2020end} (TED-LIUM~v3~\cite{tedlium3}, MSP-Podcast~\cite{mspodcast},
Earnings-21~\cite{earnings21}, Earnings-22~\cite{earnings22}) provide 
English speech; their native annotations vary by source (e.g.\ MSP-Podcast
supplies emotion, the Earnings sets supply NER), and remaining tasks are completed
by expert teachers. The multilingual read-speech corpus CML-TTS~\cite{cml} adds
German, Spanish, French, and Italian audio with native transcripts, for which
emotion, NER, and gender are expert-distilled. Every training clip thus carries
all five labels; the multilingual component broadens coverage beyond English and
counteracts the tendency of audio models to default to English output (cf.\
Section~\ref{sec:crosslingual}). Table~\ref{tab:datastats} marks, per corpus,
which labels are native and which are distilled.

\begin{table}[t]
\caption{Training and evaluation corpora: utterance counts and per-corpus
labeled tasks. Task codes: A\,=\,ASR, E\,=\,emotion, N\,=\,NER, G\,=\,gender,
L\,=\,LID. \underline{Underline} marks tasks with native (gold) labels; the remaining
tasks are filled by expert teachers. VoxPopuli
emotion is pseudo-labeled by the same teacher, whereas the native-emotion corpora
carry human (gold) emotion labels. Dutch (nl) is held out from fine-tuning and
used only for the generalization test (Section~\ref{sec:generalization}).}
\label{tab:datastats}
\centering
\footnotesize
\setlength{\tabcolsep}{4.5pt}
\renewcommand{\arraystretch}{1.05}
\begin{tabular}{llcr}
\toprule
\textbf{Corpus} & \textbf{Lang.} & \textbf{Tasks} & \textbf{\#\,Utt.} \\
\midrule
\rowcolor{rowTrain}\multicolumn{4}{l}{\textbf{Training}} \\
TED-LIUM~v3 & en & \underline{A},E,N,G,\underline{L} & 96{,}800 \\
MSP-Podcast & en & \underline{A},\underline{E},N,G,\underline{L} & 35{,}790 \\
Earnings-22 & en & \underline{A},E,\underline{N},G,\underline{L} & 22{,}511 \\
Earnings-21 & en & \underline{A},E,\underline{N},G,\underline{L} &  7{,}097 \\
CML-TTS     & de & \underline{A},E,N,G,\underline{L} & 87{,}862 \\
CML-TTS     & fr & \underline{A},E,N,G,\underline{L} & 45{,}572 \\
CML-TTS     & es & \underline{A},E,N,G,\underline{L} & 45{,}023 \\
CML-TTS     & it & \underline{A},E,N,G,\underline{L} & 43{,}340 \\
\midrule
\rowcolor{rowEval}\multicolumn{4}{l}{\textbf{Evaluation (Multitask)}} \\
VoxPopuli & en & \underline{A},E,\underline{N},\underline{G},\underline{L} & 1{,}790 \\
VoxPopuli & de & \underline{A},E,\underline{N},\underline{G},\underline{L} & 1{,}934 \\
VoxPopuli & es & \underline{A},E,\underline{N},\underline{G},\underline{L} & 1{,}478 \\
VoxPopuli & fr & \underline{A},E,\underline{N},\underline{G},\underline{L} & 1{,}621 \\
VoxPopuli & it & \underline{A},E,N,\underline{G},\underline{L} & 1{,}177 \\
VoxPopuli & nl & \underline{A},E,\underline{N},\underline{G},\underline{L} & 1{,}120 \\
\addlinespace
\rowcolor{rowEmo}\multicolumn{4}{l}{\textbf{Evaluation (Native Emotion Rec.)}} \\
IEMOCAP & en & \underline{E} & 1{,}365 \\
EmoDB   & de & \underline{E} &   107 \\
MESD    & es & \underline{E} &   204 \\
CaFE    & fr & \underline{E} &   312 \\
EMOVO   & it & \underline{E} &    98 \\
\bottomrule
\end{tabular}
\end{table}

\subsection{Expert teachers and pseudo-labeling}\label{sec:experts}
For most non-ASR tasks we apply a dedicated, publicly available model to the
source audio or its transcript and store the prediction as the training label. These are the same specialists the cascade baseline
of Fig.~\ref{fig:multimodal} chains together; we run them offline to supervise a
single joint model rather than at inference time. \emph{Emotion} labels come from
a Whisper-large-v3~\cite{radford2023whisper} classifier fine-tuned on MSP-Podcast, applied to mono $16$\,kHz segments of $3$--$15$\,s
under the nine-class MSP taxonomy. \emph{NER} spans are extracted from the
transcript by GLiNER2~\cite{gliner2} (confidence $0.5$) under a ten-type open
schema, mapped onto our nine canonical types 
and realigned to surface tokens as a $19$-tag BIO sequence. \emph{Gender} is predicted directly
from audio by a wav2vec2-XLSR-53~\cite{conneau2020unsupervised} classifier. For
\emph{LID} our pipeline includes a w2v-BERT language-identification
model~\cite{farhadipour2026subtract} trained on
VoxLingua107~\cite{valk2021voxlingua107}. In this work, however, every source corpus
carries native language tags, so we use those directly, mapped to the six-class
set. Each
completed row is validated against the closed vocabularies
(Table~\ref{tab:labels}) before serialization: rows with any out-of-vocabulary
label are discarded, and the NER training split keeps only utterances with at
least one entity after cleanup.


\section{Models \& Experimental Setup}
This section presents the systems we compare and the protocol under which they
are run. We first describe the ten open-weight baseline MLLMs, then OmniTag-LLM,
our LoRA fine-tune of Qwen3-Omni, and finally the shared prompting, decoding, and
serving setup applied uniformly to every system.

\subsection{Baseline Models}\label{sec:baselines}
We benchmark ten open-weight MLLMs covering the major omni and speech-LLM
families. Seven use dense backbones (Qwen2.5-Omni-7B~\cite{xu2025qwen25omni},
Phi-4-MM~\cite{microsoft2025phi4}, MiniCPM-o-2.6~\cite{minicpmo2024},
Kimi-Audio-7B~\cite{ding2025kimi}, Gemma-4-12B-IT~\cite{gemma},
MOSS-Audio-8B~\cite{mossaudio}, and Step-Audio-2-mini~\cite{stepaudio2}), and
three use mixture-of-experts (MoE) backbones (Gemma-4-E4B-IT~\cite{gemma},
Nemotron-Omni-30B~\cite{nemotronomni}, and Qwen3-Omni-30B~\cite{qwen3omni}). To
isolate model capability from prompt or input variation, every system receives
byte-identical prompts and audio with no task- or model-specific tuning. Serving
back-ends are detailed in Section~\ref{sec:setup}.


\subsection{OmniTag-LLM: fine-tuning Qwen3-Omni}
OmniTag-LLM is produced by LoRA fine-tuning~\cite{hu2022lora} Qwen3-Omni on the unified,
expert-distilled corpus of Section~\ref{sec:training}, under the same joint
five-field JSON objective and shared prompt (Fig.~\ref{fig:prompt}) used at
evaluation. Because every training target is written in the spoken language, the
model cannot acquire the ``transcribe-into-English'' shortcut analyzed in
Section~\ref{sec:crosslingual}.

We fine-tune Qwen3-Omni-30B-A3B-Instruct (MoE; $30$B total, $3$B active) with
LoRA via \texttt{ms-swift}~\cite{zhao2024swift}. Adapters ($r{=}8$, $\alpha{=}32$, dropout~$0.05$) are
placed on all linear layers of the thinker language model, attention
($q,k,v,o$) and MLP (gate, up, down), while the audio/vision encoders and
modality aligner stay frozen. We optimize with AdamW ($\beta_1{=}0.9$,
$\beta_2{=}0.95$, weight decay~$0.1$, gradient clipping~$1.0$) at a peak learning
rate of $1\times10^{-4}$ on a cosine schedule without warmup, effective batch
size~$12$ (per-device~$3$, accumulation~$4$, single GPU), for one epoch. Inputs
are bfloat16 without quantization, sequences capped at $2048$ tokens, and
utterances longer than $30$\,s excluded. The mixture concatenates all selected
rows once per epoch with no per-task re-weighting, preserving the naturally
English-heavy distribution.

\subsection{Experimental Setup}\label{sec:setup}
Each model is queried once per clip with the shared prompt, and its raw output is
graded by the common scorer of Section~\ref{sec:metrics}. This shared prompt pins
the output language, instructing each model to transcribe in the spoken language
rather than translate. We impose this constraint on all final results because
without it several models default to English transcripts on non-English audio,
the failure mode analyzed in Section~\ref{sec:crosslingual}. Decoding is greedy
(\texttt{temperature}{=}$0$, \texttt{max\_new\_tokens}{=}$1024$); for models known
to degenerate we apply the minimal published mitigation (e.g.\ a repetition
penalty or stop string for Kimi-Audio) and hold all decoding settings fixed
across languages, so cross-lingual differences reflect model capability, not
decoding variance. Audio is resampled to $16$\,kHz mono and never truncated;
vLLM-served models cap context at \texttt{max\_model\_len} ($8192$ tokens, $4096$
for the Qwen3-Omni thinker server).

All systems run on one workstation with
three NVIDIA RTX~PRO~6000 Blackwell Max-Q GPUs ($96$\,GB each), one model per GPU,
no tensor parallelism. All models, including both 30B models and our fine-tuned
Qwen3-Omni, run in native \texttt{bf16}/\texttt{fp16} without quantization.

\section{Results}
We organize the results in five parts. We first report per-language performance
across the six tasks, then isolate a cross-lingual transcription failure that
inflates multilingual WER, next quantify the effect of expert-distilled
fine-tuning in OmniTag-LLM, test whether those gains are genuine
rather than artifacts of teacher imitation, and finally probe generalization to a
held-out language.

\subsection{Per-language results}\label{sec:percls}
All systems produced predictions on the full $8{,}000$-clip suite, so every cell
reflects complete coverage. Table~\ref{tab:fullresults} is the main results view:
columns are models (abbreviated; full names in Section~\ref{sec:baselines}), with our
fine-tuned OmniTag-LLM in the \textbf{OmniTag} column. Per language we report six tasks: ASR, NER,
VoxPopuli emotion (Emo$_{\mathrm{VP}}$), native-emotion (Emo$_{\mathrm{SER}}$),
gender, and LID.

\begin{figure}[t]
\centering
\includegraphics[width=0.75\columnwidth]{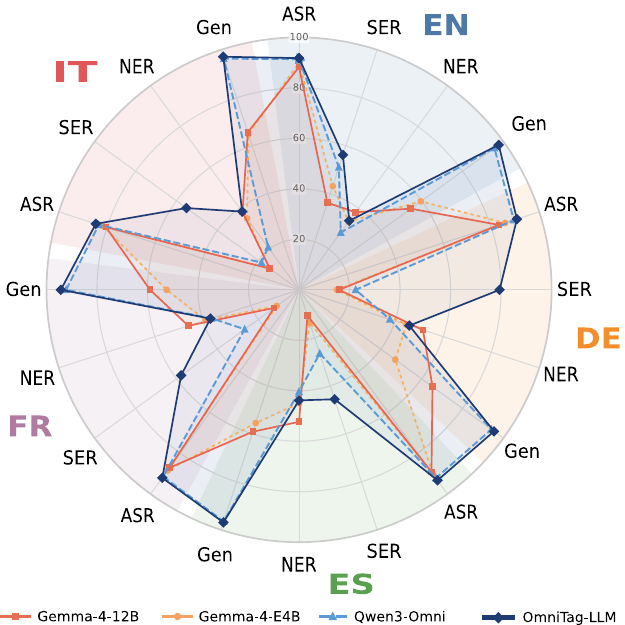}
\caption{Per-language capability profiles (unified $0$--$100$ scale; ASR shown as
$100-$WER, so farther from the center is better on every axis). Each language
sector spans ASR, native-emotion SER, entity-level NER, and gender accuracy
(labeled \emph{Gen} in the plot).
}
\label{fig:radar}
\end{figure}

Several patterns stand out. \emph{(i)} On multilingual ASR the Qwen-Omni models
lead and several compact instruction-tuned systems (Phi-4-MM, Gemma-4-E4B) are
competitive, whereas MiniCPM-o-2.6 and Nemotron-Omni-30B trail sharply on
non-English audio due to the cross-lingual failure of
Section~\ref{sec:crosslingual}. \emph{(ii)} The Gemma-4 models lead NER in four of
the five languages (Label-F1 up to $67.0$ on French; OmniTag-LLM edges ahead on
Italian), with Label-F1 consistently above Entity-F1. \emph{(iii)} Per-language
LID and gender expose joint failures that pooled scores hide: Phi-4-MM collapses
on non-English LID (e.g.\ $0.2\%$ on German), while Gemma-4 gender accuracy stays
near $47$--$66\%$ despite strong NER. Emotion is the hardest task: VoxPopuli
accuracy is only moderate ($\sim\!41$--$74\%$) and native-emotion accuracy varies
widely by corpus (e.g.\ Qwen2.5-Omni scores $70.1\%$ on EmoDB but $15.3\%$ on
EMOVO). Figure~\ref{fig:radar} visualizes these per-language profiles for the Gemma-4
family (E4B and 12B), base Qwen3-Omni, and our OmniTag-LLM.

\begin{table*}[tp]
\caption{Per-language joint-task results (\%). \textbf{Columns} = models (abbreviated names; full names in Section~\ref{sec:baselines}; rightmost \emph{shaded} column = our LoRA fine-tuned Qwen3-Omni). \textbf{Rows} group by language then task; the \textbf{Dataset} column names the evaluation corpus (Table~\ref{tab:datastats}). Paired metrics share one row as $a/b$: ASR\,=\,WER/CER ($\downarrow$), NER\,=\,Entity-F1/Label-F1, emotion\,=\,accuracy/macro-F1; gender and LID report accuracy only. All non-ASR metrics $\uparrow$. \textbf{Bold} / \underline{underline} mark the best / second-best system for \emph{each} metric.}
\label{tab:fullresults}
\centering
\footnotesize
\setlength{\tabcolsep}{2.8pt}
\renewcommand{\arraystretch}{1.05}
\setlength{\aboverulesep}{0pt}
\setlength{\belowrulesep}{0pt}
\setlength{\extrarowheight}{1pt}
\adjustbox{max width=\textwidth,max height=0.88\textheight,center}{%
\begin{tabular}{ll@{\hspace{4pt}}p{0.11\textwidth}lrrrrrrrrrr>{\columncolor{colFT}}r}
\toprule
\textbf{Lang} & \textbf{Task} & \textbf{Dataset} & \textbf{Metric} & Phi-4 & MCPM-o & Kimi-7B & MOSS-8B & Step-2 & Nemo-30B & Gemma4-4B & Gemma4-12B & Q2.5-Omni & Q3-Omni & OmniTag \\
\midrule
\multirow{6}{*}{EN} & ASR & VoxPopuli & WER/CER & \underline{7.1}/4.5 & 10.4/6.6 & 8.7/4.9 & 8.8/4.9 & 8.1/4.6 & \textbf{6.3}/\textbf{3.9} & 10.0/5.6 & 11.7/6.9 & 7.3/\underline{4.3} & 8.6/5.1 & 8.2/4.9 \\
 & NER & VoxPopuli & Ent/Lab-F1 & 6.1/11.1 & 2.3/5.6 & 7.0/11.4 & 21.1/33.3 & 12.1/19.2 & 25.9/39.5 & 33.4/\underline{48.3} & \textbf{38.0}/\textbf{50.8} & 12.8/21.2 & 28.1/38.0 & \underline{33.8}/46.0 \\
 & Emo$_{\mathrm{VP}}$ & VoxPopuli & Acc/mF1 & 69.8/15.4 & 73.6/15.7 & 72.8/10.5 & 71.7/9.6 & \textbf{74.1}/12.5 & \underline{73.9}/13.7 & 65.5/15.3 & 68.3/\underline{16.1} & 73.0/9.6 & 70.7/14.4 & 71.6/\textbf{18.5} \\
 & Emo$_{\mathrm{SER}}$ & IEMOCAP & Acc/mF1 & 25.0/13.5 & 50.3/26.1 & 44.8/15.0 & 38.6/13.3 & \textbf{56.3}/\textbf{32.7} & 37.7/16.7 & 43.2/18.3 & 36.3/13.5 & 50.1/24.9 & 51.0/19.3 & \underline{56.2}/\underline{29.1} \\
 & Gender & VoxPopuli & Acc & 67.8 & 94.4 & 81.7 & 95.0 & 82.0 & 74.7 & 59.6 & 54.7 & 91.1 & \underline{95.6} & \textbf{97.7} \\
 & LID & VoxPopuli & Acc & \underline{99.3} & 76.4 & 40.6 & 94.9 & 36.7 & 98.6 & 93.0 & 97.5 & \underline{99.3} & \textbf{99.9} & \textbf{99.9} \\
\midrule
\multirow{6}{*}{DE} & ASR & VoxPopuli & WER/CER & 13.2/8.6 & 36.5/23.8 & 21.9/12.9 & 19.9/10.8 & 22.4/12.5 & 57.7/32.4 & 14.2/8.1 & 16.8/9.7 & 11.7/7.1 & \underline{10.6}/\underline{6.9} & \textbf{9.3}/\textbf{6.1} \\
 & NER & VoxPopuli & Ent/Lab-F1 & 6.3/13.7 & 1.1/2.2 & 7.3/13.4 & 26.4/36.9 & 10.6/20.8 & 18.2/32.0 & 44.7/\underline{62.8} & \textbf{51.6}/\textbf{65.3} & 16.4/26.4 & 37.8/48.8 & \underline{45.9}/60.8 \\
 & Emo$_{\mathrm{VP}}$ & VoxPopuli & Acc/mF1 & 62.9/9.4 & 65.2/9.6 & 65.6/\underline{10.6} & 67.4/8.2 & \textbf{69.1}/9.3 & \underline{68.8}/9.2 & 57.5/8.2 & 60.7/10.3 & 68.2/7.4 & 66.0/8.7 & 66.2/\textbf{12.7} \\
 & Emo$_{\mathrm{SER}}$ & EmoDB & Acc/mF1 & 13.1/3.6 & 42.1/25.7 & 27.1/17.4 & 20.6/9.6 & 51.4/44.1 & 15.0/3.7 & 15.0/3.8 & 15.9/5.3 & \underline{70.1}/\underline{55.8} & 22.4/13.5 & \textbf{79.4}/\textbf{56.9} \\
 & Gender & VoxPopuli & Acc & 64.3 & 92.3 & 81.8 & 93.6 & 85.6 & 79.0 & 47.1 & 65.3 & 87.0 & \underline{94.1} & \textbf{95.4} \\
 & LID & VoxPopuli & Acc & 0.2 & 97.7 & 65.6 & 98.7 & \textbf{99.6} & 89.6 & 96.9 & 98.5 & \underline{99.4} & \underline{99.4} & \underline{99.4} \\
\midrule
\multirow{6}{*}{ES} & ASR & VoxPopuli & WER/CER & 8.2/5.5 & 24.2/18.3 & 10.7/6.8 & 10.9/6.4 & 12.2/7.3 & 27.8/15.6 & 9.9/6.3 & 10.8/7.0 & \underline{7.8}/\underline{5.1} & \underline{7.8}/5.3 & \textbf{6.7}/\textbf{4.1} \\
 & NER & VoxPopuli & Ent/Lab-F1 & 5.4/15.8 & 2.6/4.8 & 7.7/12.2 & 24.3/38.8 & 9.3/16.1 & 31.3/45.0 & \underline{44.9}/\underline{62.0} & \textbf{52.3}/\textbf{63.9} & 17.6/25.1 & 40.4/48.8 & 43.8/57.1 \\
 & Emo$_{\mathrm{VP}}$ & VoxPopuli & Acc/mF1 & 40.7/8.7 & 46.1/\underline{10.4} & 48.4/8.9 & 48.3/7.4 & \underline{49.0}/10.2 & 48.8/9.5 & 42.5/10.2 & 43.4/9.7 & 48.4/7.3 & 46.0/8.1 & \textbf{50.0}/\textbf{12.9} \\
 & Emo$_{\mathrm{SER}}$ & MESD & Acc/mF1 & 17.6/10.6 & 25.0/20.0 & 11.3/5.8 & 16.7/7.5 & 16.2/9.0 & 12.3/3.2 & 13.7/6.2 & 10.8/3.1 & \underline{27.0}/\underline{24.8} & 26.5/16.9 & \textbf{45.6}/\textbf{35.6} \\
 & Gender & VoxPopuli & Acc & 58.6 & 95.6 & 79.8 & 96.4 & 80.0 & 67.1 & 55.5 & 59.1 & 84.1 & \underline{96.5} & \textbf{96.9} \\
 & LID & VoxPopuli & Acc & 11.4 & 98.6 & 80.4 & 98.0 & 69.1 & 91.9 & 97.6 & 98.4 & 98.6 & \underline{99.0} & \textbf{99.1} \\
\midrule
\multirow{6}{*}{FR} & ASR & VoxPopuli & WER/CER & 10.8/7.4 & 29.4/21.2 & 14.0/8.1 & 14.3/7.9 & 14.5/8.0 & 45.4/27.5 & 11.5/6.8 & 12.9/7.8 & 9.6/6.2 & \underline{8.5}/\textbf{5.5} & \textbf{7.9}/\underline{6.1} \\
 & NER & VoxPopuli & Ent/Lab-F1 & 5.5/12.8 & 2.3/3.8 & 5.0/9.9 & 25.6/36.5 & 9.8/21.4 & 22.7/40.0 & \underline{38.9}/\underline{65.1} & \textbf{46.1}/\textbf{67.0} & 12.2/25.1 & 36.5/53.2 & 36.9/62.0 \\
 & Emo$_{\mathrm{VP}}$ & VoxPopuli & Acc/mF1 & 58.2/11.5 & 59.2/10.0 & 60.2/9.5 & 61.7/10.3 & 63.1/10.3 & \underline{63.2}/8.8 & 52.4/\textbf{13.6} & 56.0/10.4 & 61.9/7.7 & 59.1/11.5 & \textbf{65.3}/\underline{12.4} \\
 & Emo$_{\mathrm{SER}}$ & CaFE & Acc/mF1 & 11.5/7.2 & 23.7/19.9 & \underline{46.5}/35.5 & 14.1/10.7 & 40.7/\underline{41.9} & 9.0/4.3 & 10.9/6.5 & 12.2/7.7 & 14.4/12.4 & 26.6/22.0 & \textbf{57.7}/\textbf{56.1} \\
 & Gender & VoxPopuli & Acc & 61.4 & 91.6 & 75.6 & 91.4 & 77.9 & 69.3 & 52.4 & 59.0 & 87.8 & \underline{92.2} & \textbf{94.3} \\
 & LID & VoxPopuli & Acc & 9.2 & 98.6 & 70.5 & 99.3 & 98.0 & 99.6 & 95.6 & 99.7 & 80.1 & \underline{99.9} & \textbf{100.0} \\
\midrule
\multirow{6}{*}{IT} & ASR & VoxPopuli & WER/CER & 18.7/14.7 & 48.9/38.8 & 22.4/15.9 & 26.4/16.5 & 22.6/15.9 & 55.4/33.7 & 18.5/13.8 & 19.6/14.4 & 18.8/14.8 & \underline{16.6}/\underline{13.4} & \textbf{15.4}/\textbf{11.8} \\
 & NER & VoxPopuli & Ent/Lab-F1 & 3.9/13.8 & 0.6/1.4 & 6.1/13.6 & 13.3/25.8 & 7.3/15.3 & 14.0/35.0 & 35.0/\underline{53.3} & \underline{38.1}/51.4 & 13.3/27.0 & 20.8/38.3 & \textbf{38.3}/\textbf{53.7} \\
 & Emo$_{\mathrm{VP}}$ & VoxPopuli & Acc/mF1 & 51.1/\underline{10.9} & 45.8/10.1 & 53.7/8.4 & 55.2/8.4 & 55.3/10.3 & \underline{55.4}/10.2 & 46.7/10.7 & 48.7/8.6 & 54.5/7.3 & 53.4/8.1 & \textbf{56.3}/\textbf{16.6} \\
 & Emo$_{\mathrm{SER}}$ & EMOVO & Acc/mF1 & 22.4/16.3 & 19.4/12.0 & \underline{29.6}/\underline{21.3} & 14.3/3.6 & 21.4/15.4 & 14.3/3.6 & 14.3/3.2 & 14.3/3.6 & 15.3/5.5 & 18.4/10.5 & \textbf{55.1}/\textbf{51.1} \\
 & Gender & VoxPopuli & Acc & 64.0 & 95.8 & 84.9 & \underline{96.3} & 81.5 & 74.4 & 64.8 & 65.6 & 90.8 & \underline{96.3} & \textbf{97.1} \\
 & LID & VoxPopuli & Acc & 2.0 & 98.0 & 73.8 & 98.9 & 94.8 & 44.0 & 99.7 & 98.9 & \underline{99.8} & \textbf{99.9} & \textbf{99.9} \\
\bottomrule
\end{tabular}%
}
\end{table*}

\subsection{Cross-lingual transcription failure}\label{sec:crosslingual}
Under an earlier verbatim-only ASR instruction that did not pin the output
language, models were free to answer in any language. We measure how often each
model emits an \emph{English-looking} transcript on non-English audio with a
simple lexical heuristic, where a hypothesis is flagged English if it contains at least
three high-frequency English function words and more English than
source-language ones, cross-checked against the model's own \texttt{lid} field.

\begin{figure}[t]
\centering
\includegraphics[width=0.85\columnwidth]{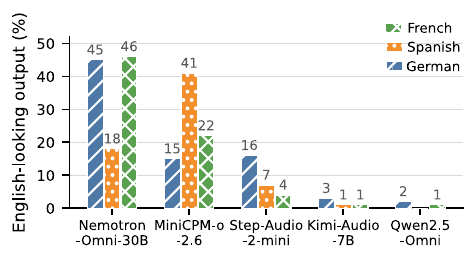}
\caption{Share of \emph{English-looking} transcripts on non-English audio under
the earlier language-unconstrained prompt (Section~\ref{sec:crosslingual}; lower
is better). }
\label{fig:crosslingual}
\end{figure}

Figure~\ref{fig:crosslingual} shows the result. Nemotron-Omni-30B and
MiniCPM-o-2.6 are most affected, emitting English on up to $45$--$46\%$ of clips
(Nemotron, German and French) and $41\%$ (MiniCPM, Spanish). Step-Audio-2-mini is
intermediate, while Kimi-Audio and Qwen2.5-Omni stay in the source language
($\leq 3\%$).

A breakdown of Nemotron-Omni-30B's German ASR makes the impact concrete. Clips
where it answers in English reach $\sim\!98$--$100\%$ WER and dominate the
$82.2\%$ corpus figure. Even its non-English-looking hypotheses still score
$\sim\!64\%$ WER, so language choice is a major but not the sole cause, since
genuine ASR quality on non-English audio is also weaker. For contrast,
Qwen2.5-Omni answers in German on $>95\%$ of clips, at $\sim\!10.6\%$ WER on that
subset.

Two practical consequences follow. First, English ASR scores must \emph{not} be
read as multilingual competence: Nemotron's $6.3\%$ English WER coexists with
German and French WER an order of magnitude higher. Second, fair multilingual
reporting should both constrain the output language in the prompt \emph{and}
report a language-matched WER. The final OmniTag-Bench prompt
(Fig.~\ref{fig:prompt}) therefore adds an explicit source-language instruction,
and we adopt the English-output rate as a standard diagnostic. Language pinning
substantially reduces the effect (Nemotron's German WER falls from $82\%$ to
$58\%$; Table~\ref{tab:fullresults}, DE/ASR) but does not eliminate it, so the
residual non-English error of Nemotron and MiniCPM still partly reflects
output-language choice rather than acoustic difficulty.

\subsection{Effect of expert-distilled fine-tuning}
Table~\ref{tab:finetune} compares OmniTag-LLM against its base Qwen3-Omni and the
strongest baselines, every entry pooled over the five languages
as in Table~\ref{tab:fullresults}. The base Qwen3-Omni-30B is already the
strongest baseline on ASR, LID, and gender, so fine-tuning chiefly targets its
principal weaknesses, structured NER and emotion, while preserving the rest.

OmniTag-LLM improves on the base model along \emph{every} axis. Entity-F1 rises
$7.0$ points ($32.7\!\rightarrow\!39.7$, $+21\%$ relative), closing more than half
the gap to the NER specialist Gemma-4-12B while retaining Qwen3-Omni's far
stronger ASR. This remaining gap should be read with scale in mind, since
Gemma-4-12B is a dense $12$B model whereas OmniTag-LLM activates only $3$B
parameters per token (of $30$B total). Against the size-comparable Gemma-4-E4B,
OmniTag-LLM is in fact competitive on NER ($39.7$ vs.\ $39.4$ Entity-F1).
WER \emph{improves} rather than regressing ($10.4\!\rightarrow\!9.5$),
confirming that joint multi-task tuning on source-language targets does not trade
away transcription, and LID stays at ceiling ($99.6\!\rightarrow\!99.7$). The
headline result is emotion: on the held-out, human-labeled
native-emotion corpora, which share \emph{no} labels and \emph{no} teacher with
our training data, accuracy nearly doubles ($29.0\!\rightarrow\!58.8$), with the
largest jumps on German EmoDB ($22.4\!\rightarrow\!79.4$) and Italian EMOVO
($18.4\!\rightarrow\!55.1$). 

\begin{table}[H]
\caption{Our fine-tuned Qwen3-Omni vs.\ its base model and the strongest
baselines on the joint objective, pooled over five languages.}
\label{tab:finetune}
\centering
\footnotesize
\setlength{\tabcolsep}{3pt}
\resizebox{0.9\columnwidth}{!}{%
\begin{tabular}{lcccccc}
\toprule
\textbf{Model} & \textbf{WER}$\downarrow$ & \textbf{Emo$_{\mathrm{VP}}$} & \textbf{Emo$_{\mathrm{SER}}$} & \textbf{Gender} & \textbf{LID} & \textbf{NER} \\
\midrule
Qwen2.5-Omni-7B      & 11.1 & 61.2 & 35.4 & 88.2 & 95.4 & 14.5 \\
Gemma-4-E4B-IT           & 12.8 & 52.9 & 19.4 & 55.9 & 96.6 & 39.4 \\
Gemma-4-12B-IT           & 14.4 & 55.4 & 17.9 & 60.7 & 98.6 & \textbf{45.2} \\
Qwen3-Omni-30B    & 10.4 & 59.1 & 29.0 & 94.9 & 99.6 & 32.7 \\
\textbf{Ours: OmniTag-LLM} & \textbf{9.5} & \textbf{61.9} & \textbf{58.8} & \textbf{96.3} & \textbf{99.7} & 39.7 \\
\bottomrule
\end{tabular}%
}
\end{table}

\subsection{Construct validity: are the gains real?}\label{sec:validity}
A legitimate concern is \emph{circularity}: since our non-ASR training labels come
from expert teachers, a fine-tune might appear to improve merely by imitating
them. This risk is confined to a single metric, because every evaluation gold but
one is teacher-independent: ASR is scored against native transcripts, NER against MSNER~\cite{meeus2024msner}, and gender and LID against VoxPopuli's native labels. Only
VoxPopuli emotion shares its teacher with training, so we anchor the
emotion claim on the held-out, human-labeled native-emotion corpora, disjoint
from both our training data and the teacher, where the near-doubling
($29.0\!\rightarrow\!58.8$) cannot be an imitation artifact; for the coupled
VoxPopuli metric we report cross-model ranking rather than absolute values.

\subsection{Generalization to an unseen language}\label{sec:generalization}
We test whether the recipe transfers beyond its five training languages by
evaluating on Dutch (nl), a language absent from fine-tuning. We use the
$1{,}120$ VoxPopuli Dutch utterances for which MSNER~\cite{meeus2024msner}
provides native NER. All labels are native except emotion, which is teacher
pseudo-labeled, and no native-emotion set is available for Dutch.
Table~\ref{tab:dutch} compares the base Qwen3-Omni with OmniTag-LLM.

Fine-tuning transfers to this unseen language on most tasks. Against the native
MSNER labels, label-level NER improves ($45.3\!\rightarrow\!53.2$), and ASR is
essentially unchanged ($13.4\!\rightarrow\!13.2$ WER). Emotion also rises on both
accuracy ($61.6\!\rightarrow\!63.0$) and macro-F1 ($9.1\!\rightarrow\!15.4$),
though these labels are teacher-generated. Gender is the exception, regressing
($93.9\!\rightarrow\!88.5$). Since Dutch is typologically close
to the German and English languages, these gains indicate that OmniTag-LLM
generalizes to related but unseen languages rather than memorizing per-language
patterns.

\begin{table}[H]
\caption{Zero-shot generalization to Dutch (nl), a language unseen during
fine-tuning.}
\label{tab:dutch}
\centering
\footnotesize
\setlength{\tabcolsep}{4pt}
\renewcommand{\arraystretch}{1.1}
\resizebox{0.8\columnwidth}{!}{%
\begin{tabular}{lcccccc}
\toprule
& \multicolumn{2}{c}{\textbf{ASR}$\downarrow$} & \textbf{NER} & \multicolumn{2}{c}{\textbf{Emotion}} & \textbf{Gender} \\
\cmidrule(lr){2-3}\cmidrule(lr){5-6}
\textbf{Model} & WER & CER & Label-F1 & Acc & mF1 & Acc \\
\midrule
Qwen3-Omni           & 13.4 & 8.6 & 45.3 & 61.6 & 9.1 & \textbf{93.9} \\
\textbf{OmniTag-LLM} & \textbf{13.2} & \textbf{8.4} & \textbf{53.2} & \textbf{63.0} & \textbf{15.4} & 88.5 \\
\bottomrule
\end{tabular}%
}
\end{table}
\FloatBarrier

\section{Discussion and Conclusion}
We presented OmniTag-Bench, a five-language benchmark for \emph{joint} speech
understanding with MLLMs; a survey of ten open systems under one protocol; and
OmniTag-LLM, an expert-distilled LoRA fine-tune of Qwen3-Omni that needs no
manual multi-task annotation. OmniTag-LLM improves on its base model across every
task and nearly doubles accuracy on held-out, human-labeled emotion corpora,
evidence of transferable rather than imitative gains, and it further extends to a
close unseen language (Dutch) without per-language fine-tuning.

Forcing one JSON object also proves discriminative, exposing failure modes
invisible to single-task tests such as format violations and truncated outputs,
and turning near-saturated LID into a cross-task consistency probe. The most
actionable finding is cross-lingual. Several strong models silently transcribe
non-English speech into English, and this, not raw acoustic difficulty, drives
much of their multilingual WER. Across systems, capability is a matter of recipe
rather than size, since instruction tuning and training-data language balance,
not parameter count, separate the strong from the weak. We release the prompt,
scorer, and labeling pipeline,\footnote{The GitHub link is removed for anonymous
review and will be added in the camera-ready version.} and recommend a joint
protocol with language-faithful prompting and language-matched WER as defaults
for multilingual MLLM evaluation.
Some limitations of the proposed system are: most non-ASR labels are
teacher-distilled, so the fine-tuning claims rest on teacher-independent metrics;
and the recipe is shown on a single backbone, leaving transfer to smaller MLLMs
open.

\FloatBarrier
\clearpage
\onecolumn

%

\makeatletter
\@IEEEsaveoriginalsectiononce
\setcounter{section}{0}%
\setcounter{subsection}{0}%
\setcounter{subsubsection}{0}%
\setcounter{paragraph}{0}%
\gdef\thesection{A}%
\gdef\thesectiondis{}%
\gdef\thesubsection{\Alph{subsection}}%
\gdef\@IEEEthmcounterinsection#1{A}%
\refstepcounter{section}%
\def\section{\@ifstar{\@IEEEappendixsavesection*}{%
                \@IEEEdestroythesectionargument}}%
\makeatother

\begin{center}
{\LARGE\bfseries Appendix\par}
\vspace{0.45em}
{\Large Supplementary Experimental Details\par}
\end{center}
\vspace{0.6em}

\noindent This appendix lists implementation details omitted from the main paper for space.

\vspace{0.4em}

\subsection{Model Roster and Inference Stack}
\label{app:models}

Table~\ref{tab:app_roster} lists the ten open-weight baseline MLLMs evaluated in
the main paper (Section~\ref{sec:baselines}). Table~\ref{tab:app_serving} records
the serving back-end, software version, and decoding notes for each system. Serving
heterogeneity can shift WER by a fraction of a point but does not affect the
order-of-magnitude effects reported in the main paper.

\begin{table}[H]
\caption{Evaluated multimodal LLM roster (10 systems). \textbf{Arch.} denotes a
dense or mixture-of-experts (MoE) language-model backbone; both 30B systems are
MoE with $\sim$3B active parameters per token.}
\label{tab:app_roster}
\centering
\footnotesize
\setlength{\tabcolsep}{5pt}
\renewcommand{\arraystretch}{1.1}
\begin{tabular}{lcc}
\toprule
\textbf{Model} & \textbf{Params} & \textbf{Arch.} \\
\midrule
Qwen2.5-Omni-7B~\cite{xu2025qwen25omni} & 7B   & Dense \\
Phi-4-MM~\cite{microsoft2025phi4}           & 5.6B & Dense \\
MiniCPM-o-2.6~\cite{minicpmo2024}           & 8B   & Dense \\
Kimi-Audio-7B~\cite{ding2025kimi}           & 7B   & Dense \\
Gemma-4-E4B-IT~\cite{gemma}                 & 4B   & Dense \\
Gemma-4-12B-IT~\cite{gemma}                 & 12B  & Dense \\
MOSS-Audio-8B~\cite{mossaudio}              & 8B   & Dense \\
Step-Audio-2-mini~\cite{stepaudio2}         & 8B   & Dense \\
Nemotron-Omni-30B~\cite{nemotronomni}       & 30B  & MoE   \\
Qwen3-Omni-30B~\cite{qwen3omni}             & 30B  & MoE   \\
\bottomrule
\end{tabular}
\end{table}

\begin{table}[H]
\caption{Inference stack per evaluated system. All models run on one workstation
with three NVIDIA RTX~PRO~6000 Blackwell Max-Q GPUs ($96$\,GB each), one model
per GPU, without tensor parallelism. Weights are loaded in native
\texttt{bf16}/\texttt{fp16} without quantization. Audio is resampled to
$16$\,kHz mono and never truncated. Decoding is greedy
(\texttt{temperature}{=}$0$, \texttt{max\_new\_tokens}{=}$1024$) unless noted.}
\label{tab:app_serving}
\centering
\footnotesize
\setlength{\tabcolsep}{3.5pt}
\renewcommand{\arraystretch}{1.08}
\begin{tabular}{@{}llp{0.58\linewidth}@{}}
\toprule
\textbf{Model} & \textbf{Back-end / version} & \textbf{Notes} \\
\midrule
Qwen2.5-Omni-7B & vLLM 0.18.0 & \texttt{max\_model\_len}{=}$8192$ \\
Phi-4-MM & vLLM 0.18.0 & speech LoRA (rank~$320$); \texttt{max\_model\_len}{=}$8192$ \\
MiniCPM-o-2.6 & vLLM 0.18.0 & \texttt{max\_model\_len}{=}$8192$ \\
Kimi-Audio-7B & vLLM 0.18.0 & repetition penalty / stop string per published recipe \\
Gemma-4-E4B-IT & vLLM nightly 0.22.1rc1 & Blackwell support; \texttt{max\_model\_len}{=}$8192$ \\
Gemma-4-12B-IT & vLLM nightly 0.22.1rc1 & Blackwell support; \texttt{max\_model\_len}{=}$8192$ \\
Nemotron-Omni-30B & vLLM nightly 0.22.1rc1 & Blackwell support; \texttt{max\_model\_len}{=}$8192$ \\
Qwen3-Omni-30B (base) & vLLM-Omni 0.22.1 & thinker-only mode; \texttt{max\_model\_len}{=}$4096$ \\
OmniTag-LLM (fine-tuned) & vLLM-Omni 0.22.1 & same stack as base Qwen3-Omni \\
Step-Audio-2-mini & StepFun vLLM fork & vendor-recommended serving path \\
MOSS-Audio-8B & HuggingFace Transformers 4.57.1 & no vLLM path available at evaluation time \\
\bottomrule
\end{tabular}
\end{table}

\subsection{OmniTag-LLM Fine-Tuning Hyperparameters}
\label{app:finetune}

Table~\ref{tab:app_finetune} consolidates the LoRA fine-tuning configuration for
OmniTag-LLM. Training uses \texttt{ms-swift} on the expert-distilled mixture of
Section~\ref{sec:training} under the shared five-field JSON prompt
(Fig.~\ref{fig:prompt}).

\begin{table}[H]
\caption{LoRA fine-tuning configuration for OmniTag-LLM (Qwen3-Omni-30B-A3B-Instruct).}
\label{tab:app_finetune}
\centering
\footnotesize
\setlength{\tabcolsep}{4pt}
\renewcommand{\arraystretch}{1.12}
\begin{tabular}{@{}ll@{}}
\toprule
\textbf{Setting} & \textbf{Value} \\
\midrule
Backbone & Qwen3-Omni-30B-A3B-Instruct (MoE; $30$B total, $3$B active) \\
Framework & \texttt{ms-swift}~\cite{zhao2024swift} \\
LoRA rank / $\alpha$ / dropout & $r{=}8$, $\alpha{=}32$, $0.05$ \\
Target modules & all linear layers of thinker LM (attn $q,k,v,o$; MLP gate/up/down) \\
Frozen modules & audio encoder, vision encoder, modality aligner \\
Optimizer & AdamW ($\beta_1{=}0.9$, $\beta_2{=}0.95$, weight decay $0.1$) \\
Learning rate & $1\times10^{-4}$, cosine schedule, no warmup \\
Gradient clipping & $1.0$ \\
Effective batch size & $12$ (per-device $3$, accumulation $4$, single GPU) \\
Epochs & $1$ \\
Precision & bfloat16, no quantization \\
Max sequence length & $2048$ tokens \\
Max utterance duration & $30$\,s (longer clips excluded) \\
Data mixing & all selected rows concatenated once per epoch, no task re-weighting \\
\bottomrule
\end{tabular}
\end{table}

\subsection{Expert Teacher Models}
\label{app:teachers}

Table~\ref{tab:app_teachers} lists the specialist models used to distil non-ASR
training labels (Section~\ref{sec:experts}). Transcripts and language identity are
always taken from native corpus references or metadata and are never
machine-generated.

\begin{table}[H]
\caption{Expert teacher models used to distil the non-ASR training labels.}
\label{tab:app_teachers}
\centering
\scriptsize
\begin{tabular}{@{}llp{0.46\columnwidth}@{}}
\toprule
\textbf{Task} & \textbf{Teacher model} & \textbf{Output / notes} \\
\midrule
Emotion & Whisper-large-v3 (MSP fine-tune) & 9-class MSP taxonomy; mono $16$\,kHz segments of $3$--$15$\,s \\
NER & GLiNER2-multi-v1 & span extraction at confidence $0.5$; mapped to 9 canonical types and 19-tag BIO \\
Gender & wav2vec2-XLSR-53 classifier & male / female from audio \\
LID & native corpus metadata & mapped to 6-class prompt vocabulary \\
ASR & native transcript references & verbatim text after markup cleanup \\
\bottomrule
\end{tabular}
\end{table}

For NER training rows we keep only utterances with at least one entity after
cleanup. Each completed row is validated against the closed vocabularies of
Table~\ref{tab:labels}; rows with any out-of-vocabulary label are discarded.

\subsection{Dataset Preparation}
\label{app:datasets}

This subsection summarizes what was done for each corpus in
Table~\ref{tab:datastats}. Task codes follow the main paper: A\,=\,ASR,
E\,=\,emotion, N\,=\,NER, G\,=\,gender, L\,=\,LID. Underlined tasks in
Table~\ref{tab:datastats} are native; remaining tasks are filled by the teachers
of Appendix~\ref{app:teachers}.

\subsubsection{Training corpora}

\textbf{TED-LIUM~v3} (en, 96{,}800 utt.). We use the released word-level speech
segments with native transcripts (A, L). Remaining labels (E, N, G) are
teacher-distilled offline. Transcripts are normalized by removing non-speech markup
before teacher calls and JSON serialization.

\textbf{MSP-Podcast} (en, 35{,}790 utt.). Native audio, transcript, and emotion
labels (A, E, L) are taken from the corpus; NER and gender are teacher-distilled.
Emotion labels are mapped onto the nine-class evaluation taxonomy
(Table~\ref{tab:labels}).

\textbf{Earnings-21 / Earnings-22} (en, 7{,}097 / 22{,}511 utt.). These earnings-call
sets supply native transcripts and NER annotations (A, N, L). Emotion and gender
are teacher-distilled. We retain the official train partitions used in prior
spoken-NER work and serialize each clip into the unified JSON schema.

\textbf{CML-TTS} (de/fr/es/it, 43{,}340--87{,}862 utt. per language). Read speech
with native transcripts and language tags (A, L). Emotion, NER, and gender are
teacher-distilled for every clip so that all five tasks can supervise joint
training. The multilingual component is included deliberately to reduce
English-default transcription behaviour (Section~\ref{sec:crosslingual}).

\subsubsection{Evaluation corpora}

\textbf{VoxPopuli multitask splits} (en/de/es/fr/it/nl). We sample fixed per-language
evaluation subsets (Table~\ref{tab:datastats}) from VoxPopuli~\cite{voxpopuli}.
Transcript, gender, and LID are native. NER uses MSNER~\cite{meeus2024msner} for
all languages except Italian, where MSNER is unavailable and NER is pseudo-labeled
with GLiNER2. Emotion on VoxPopuli is pseudo-labeled with the same Whisper MSP
teacher used in training and should be read as a relative comparison
(Section~\ref{sec:validity}).

\textbf{Native emotion corpora} (IEMOCAP, EmoDB, MESD, CaFE, EMOVO). One acted
emotion set per language receives the same joint five-task prompt as VoxPopuli,
but only emotion accuracy and macro-F1 are scored because no ground truth exists
for the other four tasks on these sets.

\textbf{Dutch generalization split} (nl, 1{,}120 utt.). Dutch is excluded from
fine-tuning and used only for the unseen-language test in
Section~\ref{sec:generalization}. Labels follow the same construction as the other
VoxPopuli multitask splits.

\subsection{Scoring and Parsing Details}
\label{app:scoring}

All models are graded with one shared scorer (Section~\ref{sec:metrics}). JSON
responses are parsed leniently: single quotes, trailing commas, and array-wrapped
objects are accepted. A plain-text response is treated as transcript-only so ASR
remains scorable when the JSON instruction is ignored; an unparseable response
counts as incorrect for all non-ASR tasks.

ASR hypotheses and references undergo Unicode NFKC normalization, lowercasing, and
punctuation removal before WER/CER computation. NER spans are extracted as
\texttt{(phrase, type)} pairs and scored with Entity-F1 (strict surface match) and
Label-F1 (type match only), both computed over $100\%$ of clips.

Because several native-emotion evaluation sets are small (EMOVO: $98$ clips), we
treat sub-point differences cautiously. For headline comparisons we recommend
$95\%$ bootstrap confidence intervals with clip-level resampling ($10{,}000$
replicates).

\subsection{Extended Limitations}
\label{app:limitations}

\textbf{Distillation coupling.} Non-ASR \emph{training} labels inherit teacher
biases, and VoxPopuli \emph{emotion} shares the training teacher; its absolute
scores should be read as relative comparisons. All other evaluation labels used in
Section~\ref{sec:validity} are teacher-independent.

\textbf{Single run and small splits.} Reported results are single-run greedy
decodes without confidence intervals unless bootstrap intervals are computed
separately.

\textbf{Serving heterogeneity.} Back-ends differ across models
(Appendix~\ref{app:models}); differences on the order of tenths of a WER point
between systems are not meaningful.

\textbf{Single prompt.} The fixed instruction isolates multilingual and multi-task
ability from instruction variety; prompt robustness is not measured here.

\textbf{Cross-lingual diagnostic.} The English-output detector used in
Section~\ref{sec:crosslingual} is lexical and approximate; it is intended for
aggregate trends, not per-clip judgements.

\textbf{Task framing.} Gender is a binary perceptual-acoustic label inherited from
teachers and source corpora; we frame it as speaker-attribute estimation for
downstream indexing, not identity inference.

\clearpage
\twocolumn

\bibliographystyle{IEEEtran}
\bibliography{references}

\end{document}